\pdfoutput=1

\documentclass[11pt]{article}

\usepackage[]{ACL2023}

\usepackage{times}
\usepackage{latexsym}
\usepackage{booktabs}

\usepackage[T1]{fontenc}

\usepackage[utf8]{inputenc}

\usepackage{microtype}

\usepackage{inconsolata}

\usepackage{graphicx}
\usepackage{xcolor}
\usepackage{enumitem}
\usepackage{hyperref}

\usepackage{todonotes}

\newcommand{\strict}[1]{\textsc{Strict}}
\newcommand{\strictsmall}[1]{\textsc{Strict-small}}
\newcommand{\paper}[1]{\textsc{Paper}}
\newcommand{\vision}[1]{\textsc{Vision}}

\title{[Call for Papers] The 2\textsuperscript{nd} BabyLM Challenge: Sample-efficient pretraining\\on a developmentally plausible corpus  \\ 
 \vspace{0.2cm} \url{https://babylm.github.io/}}

\author{
        Leshem Choshen \\ IBM Research, MIT
        \And
        Ryan Cotterell \\ ETH Zürich
        \And
        Michael Y. Hu \\ NYU
        \And
        Tal Linzen \\ NYU
        \And
        Aaron Mueller \\ Northeastern University
        \AND
        Candace Ross \\ Meta AI
        \And
        Alex Warstadt \\ ETH Zürich
        \And
        Ethan Wilcox \\ ETH Zürich
        \And
        Adina Williams \\ Meta AI
        \And
        Chengxu Zhuang \\ MIT
}

\begin{document}

\maketitle

\begin{abstract}

After last year's successful BabyLM Challenge, the competition will be hosted again in 2024/2025. The overarching goals of the challenge remain the same; however, some of the competition rules will be different. The big changes for this year's competition are as follows: First, we replace the loose track with a \emph{paper track}, which allows (for example) non-model-based submissions, novel cognitively-inspired benchmarks, or analysis techniques. Second, we are relaxing the rules around pretraining data, and will now \emph{allow participants to construct their own datasets} provided they stay within the 100M-word or 10M-word budget. Third, we introduce a \emph{multimodal vision-and-language track}, and will release a corpus of 50\% text-only and 50\% image--text multimodal data as a starting point for LM model training. The purpose of this CfP is to provide rules for this year's challenge, explain these rule changes and their rationale in greater detail, give a timeline of this year's competition, and provide answers to frequently asked questions from last year's challenge.

\end{abstract}

\section{The BabyLM Challenge}

After last year's successful BabyLM Challenge, the competition will be hosted again in 2024/2025. The overarching goals of the challenge remain the same, however, some of the competition rules will be different. The purpose of this CFP is to provide a timeline of this year's competition, explain the rule changes and their rationale, and provide answers to frequently asked questions from last year's challenge.

As before, the goal of the BabyLM shared task is to incentivize researchers interested in pretraining and/or cognitive modeling to focus their efforts on optimizing pretraining given data limitations inspired by human development. Additionally, we hope to democratize research on pertaining by drawing attention to open problems that can be addressed on a university budget. Please see the previous call for paper \citep{warstadt2023call} or the introduction to last year's proceedings \citep{warstadt2023findings} for a deeper discussion of the rationale behind the challenge.

This year's version of the BabyLM challenge will remain the same in terms of goals, but with a few key differences, which we list below:
\begin{itemize}[leftmargin=0.5cm,rightmargin=0.5cm]

    \item To encourage contributions that are related to the goals of the challenge, but do not involve direct competition entries, \textbf{we are introducing a paper-only track.} Paper track submissions could include things like novel cognitively-inspired evaluation metrics or in-depth analyses of one particular BabyLM model.

    \item Last year, all competition entrants were required to pre-train on a fixed corpus, meaning that submissions could not investigate the impact of data quality on pretraining. This year we will relax this requirement. While we will still provide language-only and multi-modal (see below) datasets of 100M and 10M words, \textbf{participants are free to construct their own datasets, provided that they stay within the 100M or 10M word budget.} Participants will be asked to provide a datasheet for any datasets they construct themselves.

    \item Human language learning is inherently multi-modal. To encourage more multi-modal submissions, \textbf{we are replacing last year's loose track with a vision-language track}. For this track, we will release a corpus of 50\% text-only and 50\% image--text multimodal data.
    
\end{itemize}

\section{Key Dates}

\begin{itemize}[leftmargin=0.5cm,rightmargin=0.5cm,itemsep=0em]
    \item \textbf{March 30 2024:} Training data released
    \item \textbf{April 30 2024:} Evaluation pipeline released
    \item \textbf{September 13 2024:} Results due
    \item \textbf{TBD:} Papers due
    \item \textbf{TBD:} Peer review begins
    \item \textbf{TBD:} Peer review ends, acceptances and leaderboard released
    \item \textbf{TBD:} Presentations
\end{itemize}

\begin{table*}[t]
    \centering
    \resizebox{\linewidth}{!}{
    \begin{tabular}{llrrr}
    \toprule
    Dataset & Description & \# Words (multimodal track) & \# Words (strict track) & \# Images\\
    \midrule
    Localized Narratives \cite{LocalizedNarratives} & Image Caption & 27M & -- & 0.6M \\
    Conceptual Captions 3M \cite{CC3M} & Image Caption & 23M & -- & 2.3M \\
    CHILDES \citep{macwhinney2000childes} & Child-directed speech & 15M & 29M & -- \\
    British National Corpus (BNC), dialogue portion & Dialogue & 4M & 8M & -- \\
    Project Gutenberg (children's stories) \citep{gerlach-2018-gutenberg} & Written English & 13M & 26M & -- \\
    OpenSubtitles \citep{lison-tiedemann-2016-opensubtitles2016} & Movie subtitles & 10M & 20M & -- \\
    Simple English Wikipedia & Written Simple English & 7M & 15M & -- \\
    Switchboard Dialog Act Corpus \citep{Stolcke-etal:2000} & Dialogue & $<$1M & 1M & -- \\
    \midrule
    \emph{Total} & -- & 100M & 100M & 2.9M\\
    \bottomrule
    \end{tabular}}
    \caption{Datasets for the multimodal and strict tracks of the 2nd BabyLM Challenge. Word counts are approximate and subject to slight changes. 
    \looseness=-1}
    \label{tab:data}
\end{table*}

\section{Participation Rules: Tracks}

The second BabyLM Challenge includes three tracks: \textbf{\strict}, \textbf{\strictsmall}, and \textbf{\vision} and \textbf{\paper}. 

The \strict{} and \strictsmall{} tracks require that submissions be trained on 100M words or less (in \strict{}) and 10M words or less (in \strictsmall{}). These tracks no longer require that participants use the fixed dataset from last year's challenge, although we will still provide an updated version of this dataset for participants interested in using it. See Section \ref{sec:dataset} for details on the updated version of this language-only dataset. Models in these tracks will be evaluated on language-only evaluation tasks.

In the \vision{} track, participants will train multi-modal image-text models. Participants can use whatever training procedure they wish, as long as models can provide (pseudo) log-likelihoods to strings of text, conditioned on an image. Participants are free to use whatever data they wish, as long as the dataset is within a 100M word budget. To facilitate easier participation in this track, we will release a suggested multimodal dataset that consists of 50\% text-only and 50\% paired image--text data. Submissions to this track will be evaluated on language-only tasks, as well as multi-modal tasks.

In the \paper{} track, participants are invited to submit a paper on a topic relevant to cognitive modeling with language models and scaled-down pertaining. Some submissions we have in mind include novel cognitively-inspired evaluation metrics or an in-depth analysis of one particular BabyLM model. Submissions to this track are welcome to include a BabyLM model, and to report its scores, however, evaluation task scores will not necessarily determine the best-paper awardee from this track.

\section{Participation Rules: Provided Dataset}\label{sec:dataset}

For this year, we update the dataset from the text-only dataset from the previous year and provide a novel multi-modal vision/text dataset for the \vision{} track. Note that, as long as participants stay under the 100M or 10M word budget, they are free to create their own datasets instead of using the ones provided. Data for both the multimodal and text-only datasets can be downloaded from \url{https://osf.io/ad7qg/}

For the text-only dataset updates, we replace the QED portion of the previous dataset with data from CHILDES. This was done because, on further inspection, it was found that the QED dataset was of poorer quality than hoped for. CHILDES now comprises a larger portion of the new dataset, but the relative proportions of the rest of the data sources remain the same. Please see the previous year's CFP for more details about these data sources \citep{warstadt2023call}

In addition, we provide a novel image--text dataset to facilitate easier participation in the \vision track. This dataset includes two components: First, we provide 50M words of text-only data, which is downsampled from the 100M corpus (that is, we preserve the relative distribution from the different data sources). Second, we provide paired text--image data that includes 50M words of text. This dataset is drawn from two sources: 27M words from the Localized Narratives \citep{pont2020connecting} and 23M words from the conceptual captions 3M dataset \citep{sharma2018conceptual}.
For the Localized Narratives dataset, our challenge uses the text captions and the images from the MS-COCO~\cite{MSCOCO} and Open Images~\cite{OpenImages} subset in this dataset.
For the CC3M dataset, we used the image-caption pairs whose images were still valid in January 2024.
In the OSF directory at the above link, we provide scripts to download the images.
We will provide visual embeddings of these images computed from a visual encoder pretrained using DINOv2~\cite{DINOv2}, one of the SOTA unsupervised learning algorithms.
In addition to the text-image data, we provide half of the 100 million words from the text-only dataset as a text-only portion of this dataset.
We take the first half of every corpus within the text-only corpora mentioned above.
Table \ref{tab:data} gives an overview of the datasets used.

If not using our pre-fixed datasets, participants will be required to submit a datasheet for their dataset \citep{gebru2021datasheets}.

\section{Participation Rules: Evaluation}

As in the previous year, we will distribute a shared evaluation pipeline based on Google Colab. For the \strict{} and \strictsmall{} track, evaluation tasks will be, largely, the same as the previous year. For the \vision{} track, we will include an additional set of multimodal evaluation tasks. As before, our evaluation code will be public, so that those who wish to use their own computational resources instead of the Google Colab may do so. More details about the evaluation pipeline and the set of tasks will be released subsequently.

In response to comments from the previous year, this year's evaluation pipeline will use \texttt{catwalk}, which will make it easier to submit models that do not conform to HuggingFace \texttt{transformers} library (although it will still be easy to submit models if they \emph{do} conform to HuggingFace standards). All models must be able to assign a (pseudo) log-likelihood to a string of text and must be able to be fine-tuned to perform classification tasks. Models do not need to be able to generate sequences. Submissions must include model outputs for each of the core evaluations in a format that we specify in our evaluation pipeline.

\subsection{Baselines}

We will release a series of baseline models. As opposed to last year's baselines, which were trained relatively naively, this year's baseline will be based on the winning submissions from last year. For the \strict{} and \strictsmall{} tracks, we will release the following baselines:  GPT2 (decoder-only; \citealp{radford2019language}), LTG-Bert (encoder-only; \citealp{LTGBert}), and Contextualizer (decoder-only; \citealp{Contextualizer}). 
For the \vision{} track, we will release the GIT~\cite{wang2022git} and Flamingo~\cite{alayrac2022flamingo} baselines.
These baselines are meant to encourage participants to innovate and improve beyond existing models.

\section{The Submission Process}

\subsection{Results submission}

As with last year, submissions will be made using the Dynabench platform, which is an online platform for dynamic data collection and model benchmarking.\footnote{\url{https://dynabench.org/}}

\vspace{0.2cm}
\noindent\textbf{What you Need to Submit:}
\vspace{-0.3cm}
\begin{itemize}[leftmargin=0.5cm,rightmargin=0.5cm, itemsep=0cm]
    \item A link where we can download the model (any file-hosting service will do)
    \item Model predictions for our evaluation pipeline
    \item A datasheet for your dataset and a download link (if not using ours)
\end{itemize}
\vspace{-0.3cm}

\subsection{Paper submission}

Submissions will be made through OpenReview. All submissions will be full archival papers and can be up to eight pages in length. Formatting requirements for submissions will be announced at a later date, once the presentation venue has been finalized. As before, we do allow dual-submission, however, we do not allow dual publication.

\subsection{Review \& Publication}

BabyLM will hold its own review process. The acceptance criteria are based on soundness: We plan only to reject submissions that make incorrect or unjustified claims. Other feedback will be directed at the improvement of submissions.

\section{FAQs}

\paragraph{Can papers be submitted to multiple tracks?} 
Yes. For example, a single paper can describe models that are submitted separately to the \strict{} and \vision{} tracks. 

\paragraph{Can I submit a paper about my work?}
Yes, we require that all submissions be accompanied by a paper, which can be up to eight pages in length (although it doesn't have to be). Papers will be published in an archival format. All papers can describe experiments and analyses beyond the scope of the competition, and this type of non-competition contribution is required for the \paper{} track.

\paragraph{Can I submit additional evaluation metrics?}
Yes, you may submit additional evaluation metrics alongside a competition model in the \strict{}, \strictsmall{}, and \vision{} tracks. This type of contribution is especially encouraged in the \paper{} track.

\paragraph{What training regimes are permitted?}
Any kind of training objective/regime is permitted, as long as the data restrictions are followed. If you use ancillary models, for example for reranking or data augmentation, then the training data for these models is counted towards your 100M word budget.

For evaluation purposes, we require that the model provides a function to score a sequence of words without the need for additional fine-tuning.

\paragraph{Are there any limits on hyperparameters?}
No. But please share at the end what you found so we can learn from your efforts.

\paragraph{Are there any limits on the number of epochs?}
No. We put no restrictions on the number of epochs, for several reasons: First, from an engineering perspective, training LMs with SGD tends to require multiple epochs at these scales to achieve peak performance. Second, from a cognitive perspective, humans have a memory of linguistic experience and can continue to access and learn from these memories. Third, we try not to make a stand on implementations to allow the most freedom for innovation. Our internal results suggest, however, that under regular circumstances, over-training on more than a couple epochs give minor gains at most. 

\paragraph{Can I use external tools?}
Yes, but if they learned on language their tokens are counted towards the 100M. That means one can train on the same text both a tokenizer, a parser, an LM etc. or on parts of the 100M, but the sum of all text seen by all training can not surpass the amount of text allowed. This raises the questions of synthetic data. It is allowed, under some restrictions. You may generate the 100M tokens in any legal way you like (yes, distilling or writing your own is fair, if you figure what text facilitates learning it is interesting regardless of how to gather such text), you may also train eventually on more than 100M words by augmentation, however that only works in a closed system, i.e., the augmentors training data counts toward the limit, so for example training two LMs on half of the words, and then have them generate more words and train a model on both the original data and the new one is legit (and it was not tested in the previous competition, so even the example itself is interesting).

\paragraph{I have different modalities that can help}
If it is not linguistic data, prove it, last year's submission did not gain from non-linguistic grounding, but we encourage such scientific questions. If it is linguistic in nature (e.g., audio), then the words should still count towards the overall number of learnt words.

\section{Organizing Committee}
(Alphabetical by last name) Leshem Choshen, Ryan Cotterell, Michael Hu, Tal Linzen, Aaron Muller, Candace Ross, Alex Warstadt, Ethan Wilcox, Adina Williams, Chengzu Zhuang. Feel free to contact members of the organizing committee at: \texttt{leshem.choshen@mail.huji.ac.il}, \texttt{aa.mueller@northeastern.edu}, \texttt{alexwarstadt@gmail.com}, \texttt{chengxuz@mit.edu}

\bibliography{custom}
\bibliographystyle{acl_natbib}

\end{document}